\RequirePackage{fix-cm}
\documentclass[twocolumn]{svjour3}          
\smartqed  
\usepackage{graphicx}
\usepackage[utf8]{inputenc}
\usepackage[english]{babel}
\usepackage{makeidx}
\usepackage{graphicx}
\usepackage[hidelinks]{hyperref}
\usepackage{amsmath}
\usepackage[inline]{enumitem}
\usepackage{multirow}
\usepackage{subcaption}
\usepackage{caption}
\usepackage{color}
\usepackage{booktabs}
\usepackage{siunitx}
\usepackage{cite}
\usepackage{algpseudocode}
\usepackage{algorithmicx}
\usepackage{array}
\usepackage{float}
\usepackage{todonotes}
\usepackage[keeplastbox]{flushend}
\usepackage[linesnumbered,ruled,vlined]{algorithm2e}
\usepackage{amsfonts}
\usepackage{threeparttable}
\usepackage{ragged2e}
\usepackage[normalem]{ulem}
\journalname{Neural Computing and Applications}

\newcommand{\RV}[1]{{\color{black}{#1}}}

\begin{document}

\title{Tensor Pooling Driven Instance Segmentation Framework for Baggage Threat Recognition}


\author{Taimur Hassan\textsuperscript{$\star$}         \and
        Samet Ak\c{c}ay \and
        Mohammed Bennamoun \and
        Salman Khan \and
        Naoufel Werghi
}


\institute{T. Hassan (\textsuperscript{$\star$}Corresponding Author) \at
              Center for Cyber-Physical Systems (C2PS), Department of Electrical Engineering and Computer Sciences, Khalifa University, Abu Dhabi, United Arab Emirates. \\
              \email{taimur.hassan@ku.ac.ae}           
           \and
           S. Ak\c{c}ay \at
              Department of Computer Sciences, Durham University, Durham, United Kingdom.
              \and
              M. Bennamoun \at Department of Computer Science and Software Engineering, The University of Western Australia, Perth, Australia.
              \and
              S. Khan \at Mohamed bin Zayed University of Artificial Intelligence, Abu Dhabi, United Arab Emirates.
              \and
              N. Werghi \at Center for Cyber-Physical Systems (C2PS), Department of Electrical Engineering and Computer Sciences, Khalifa University, Abu Dhabi, United Arab Emirates.
}

\date{Received: date / Accepted: date}

\maketitle

\begin{abstract}
Automated systems designed for screening contraband items from the X-ray imagery are still facing difficulties with high clutter, concealment, and extreme occlusion. In this paper, we addressed this challenge using a  novel multi-scale contour instance segmentation framework that effectively identifies the cluttered contraband data within the baggage X-ray scans. Unlike standard models that employ region-based or keypoint-based techniques to generate multiple boxes around objects, we propose to derive proposals according to the hierarchy of the regions defined by the contours. The proposed framework is rigorously validated on three public datasets, dubbed GDXray, SIXray, and OPIXray, where it outperforms the state-of-the-art methods by achieving the mean average precision score of 0.9779, 0.9614, and 0.8396, respectively. Furthermore, to the best of our knowledge, this is the first contour instance segmentation framework that leverages multi-scale information to recognize cluttered and concealed contraband data from the colored and grayscale security X-ray imagery.


\end{abstract}

\section{Introduction}
X-ray imagery is a widely used modality for non-destructive testing \cite{Tang2020TII}, especially for screening illegal and smuggled items at airports, cargoes, and malls. Manual baggage inspection is a tiring task and susceptible to errors caused due to exhausting work routines and less experienced personnel.  Initial systems proposed to address these problems employed conventional machine learning \cite{bastan2013BMVC}. Driven by hand-engineered features, these methods are only applicable to limited data and confined environmental settings \cite{turcsany2013improving}.  Recently, attention has turned to deep learning methods, which gave a neat boost in accuracy and generalization capacity towards screening prohibited baggage items \cite{Hu2020ACCV, akccay2016transfer}. However, deep learning methods are also prone to clutter, and occlusion \cite{akcay2018using}. This limitation emanates from the proposal generation strategies which have been designed for the color images \cite{gaus2019evaluation}.  Unlike RGB scans, X-ray imagery lack texture and exhibit low-intensity variations between cluttered objects. This intrinsic difference makes the region-based or anchor-based proposal generation methods such as Mask R-CNN \cite{maskrcnn}, Faster R-CNN \cite{fasterrcnn}, RetinaNet \cite{retinanet}, and YOLO \cite{yolov3} less robust for detecting the cluttered contraband data \cite{akcay2018using}. Moreover, the problem is further accentuated by the class imbalance nature of the contraband items in the real-world \cite{gaus2019evaluation}. 
Despite the considerate strategies proposed to alleviate the occlusion and the imbalance nature \cite{opixray, miao2019sixray}, recognizing threatening objects in highly cluttered and concealed scenarios is still an open problem \cite{ackay2020}.

\RV{
\subsection{Contributions}
\noindent In this paper, we propose a novel multi-scale contour instance segmentation framework for identifying suspicious items using X-ray scans. Unlike standard models that employ region-based or keypoint-based techniques to generate multiple boxes around objects \cite{akccay2016transfer, akcay2018using, gaus2019evaluating}, we propose to derive proposals according to the hierarchy of the regions defined by the contours.  
The insight driving this approach is that contours are the most reliable cue in the X-ray scans due to the lack of surface texture. For example, the occluded items exhibit different transitional patterns based upon their orientation, contrast, and intensity. We try to amplify and exploit this information through the multi-scale scan decomposition, which boosts the proposed framework's capacity for detecting the underlying contraband data in the presence of clutter. Furthermore, we are also motivated by the fact that organic material's suspicious items show only their outlines in the X-ray scans \cite{hassan2019}.  
To summarize, the main features of this paper are:

\begin{itemize}[leftmargin=*]
\item Detection of overlapping suspicious items by analyzing their predominant orientations across multiple scales within the candidate scan. Unlike \cite{hassan2019, Hassan2020ACCV, hassan2020Sensors}, we propose a novel tensor pooling strategy to decompose the scan across various scales and fuses them via a  single multi-scale tensor. This scheme results in more salient contour maps (see Figure \ref{fig:fig1}), boosting our framework's capacity for handling dulled, concealed, and overlapping items.

\item  A thorough validation on three publicly available large-scale baggage X-ray datasets, including the OPIXray \cite{opixray}, which is the only dataset allowing a quantitative measure of the level of occlusion.

\item  Unlike state-of-the-art methods such as CST \cite{hassan2019}, TST \cite{Hassan2020ACCV}, and DTS \cite{hassan2020Sensors}, the performance of the proposed framework to detect occluded items has been quantitatively evaluated on \RV{OPIXray \cite{opixray} dataset}. Please see Table \ref{tab:tab5} for more details.
\end{itemize}
}

\section{Related Work} \label{sec:related}
\noindent 
Many researchers have developed computer-aided screening systems to identify potential baggage threats \cite{mery2016}. While a majority of these frameworks are based on conventional machine learning \cite{bastan2013object}, the recent works also employ supervised \cite{akccay2016transfer}, and unsupervised \cite{akccay2019skip} deep learning, and these methods outperform conventional approaches both in terms of performance, and efficiency \cite{akcay2018using}. In this section, we discuss some of the major baggage threat detection works. We refer the readers to \cite{Mery2017TMSC,ackay2020} for an exhaustive survey.

\subsection{Traditional Methods}
\noindent The early baggage screening systems were driven via classification \cite{turcsany2013improving}, segmentation \cite{heitz2010} and detection \cite{bastan2015} approaches to identify potential threats and smuggled items. Here, the work of Bastan et al. \cite{bastan2013BMVC} is appreciable, which identifies the suspicious and illegal items within the multi-view X-ray imagery through fused SIFT and SPIN driven SVM model. Similarly, SURF \cite{heitz2010}, and FAST-SURF \cite{kundegorski2016} have also been used with the Bag of Words \cite{bastan2011} to identify threatening items from the security X-ray imagery. Moreover, approaches like adapted implicit shape model \cite{riffo2015automated} and adaptive sparse representation \cite{mery2016} were also commendable for screening suspicious objects from the X-ray scans.  

\subsection{Deep Learning Frameworks}
\noindent The deep learning-based baggage screening frameworks have been broadly categorized into supervised and unsupervised \RV{learning schemes.} 

\subsubsection{Supervised Methods}
The initial deep learning approaches involved scan-level classification to identify the suspicious baggage content \cite{akccay2016transfer}. However, with the recent advancements in object detection, researchers also employed sophisticated detectors like RetinaNet \cite{retinanet}, YOLO \cite{yolo, yolov2}, and Faster R-CNN \cite{fasterrcnn} to not only recognize the contraband items from the baggage X-ray scans but also to localize them via bounding boxes \cite{akcay2018using}. Moreover, researchers also proposed semantic segmentation \cite{an2019} and instance segmentation \cite{Hassan2020ACCV} models to recognize threatening and smuggled items from the grayscale and colored X-ray imagery. Apart from this, Xiao et al. \cite{_45} presented an efficient implementation of Faster R-CNN \cite{fasterrcnn} to detect suspicious data from the TeraHertz imagery. Dhiraj et al. \cite{_42} used Faster R-CNN \cite{fasterrcnn}, YOLOv2 \cite{yolov2}, and Tiny YOLO \cite{yolov2} to screen baggage threats contained within the scans of a publicly available GDXray dataset \cite{mery2015gdxray}. Gaus et al. \cite{gaus2019evaluation} utilized RetinaNet \cite{retinanet}, Faster R-CNN \cite{fasterrcnn}, Mask R-CNN \cite{maskrcnn} (driven through ResNets \cite{he2016deep}, VGG-16 \cite{vgg16}, and SqueezeNet \cite{i2016squeezenet}) to detect prohibited baggage items. In another approach \cite{gaus2019evaluating}, they analyzed the transferability of these models on a similarly styled X-ray imagery contained within their local dataset as well as the SIXray10 subset of the publicly available SIXray dataset \cite{miao2019sixray}. Similarly, Ak\c{c}ay et al. \cite{akcay2018using} compared Faster R-CNN \cite{fasterrcnn}, YOLOv2 \cite{yolov2}, R-FCN \cite{rfcn}, and sliding-window CNN with the AlexNet \cite{alexnet} driven SVM model to recognize occluded contraband items from the X-ray imagery. Miao et al. \cite{miao2019sixray} explored the imbalanced nature of the contraband items in the real-world by developing a class-balanced hierarchical refinement (CHR) framework. Furthermore, they extensively tested their framework (backboned through different classification models) on their publicly released SIXray \cite{miao2019sixray} dataset. Wei et al. \cite{opixray} presented a plug-and-play module dubbed De-occlusion Attention Module (DOAM) that can be coupled with any object detector to enhance its capacity towards screening occluded contraband items. DOAM was validated on the publicly available OPIXray \cite{opixray} dataset, which is the first of its kind in providing quantitative assessments of baggage screening frameworks under low, partial, and full occlusion \cite{opixray}. \RV{
Apart from this, Hassan et al. \cite{hassan2019} also addressed the imbalanced nature of the contraband data by developing the cascaded structure tensors (CST) based baggage threat detector. CST \cite{hassan2019} generates a balanced set of contour-based proposals, which are then utilized in training the backbone model to screen the normal and abnormal baggage items within the candidate scan \cite{hassan2019}. Similarly, to overcome the need to train the threat detection systems on large-scale and well-annotated data, Hassan et al. \cite{hassan2020Sensors} introduced meta-transfer learning-based dual tensor-shot (DTS) detector. DTS \cite{hassan2020Sensors} analyzes the scan's saliency to produce low and high-density contour maps from which the suspicious contraband items are identified effectively with few-shot training \cite{hassan2020Sensors}.  In another approach, Hassan et al. \cite{Hassan2020ACCV} developed an instance segmentation-based threat detection framework that filters the contours of the suspicious items from the regular content via trainable structure tensors (TST) \cite{Hassan2020ACCV} to identify them accurately within the security X-ray imagery. 
}
\subsubsection{Unsupervised Methods}
\noindent While most baggage screening frameworks involved supervised learning, researchers have also explored adversarial learning to screen contraband data as anomalies. Ak\c{c}ay et al. \cite{akcay2018ganomaly}, among others, laid the foundation of unsupervised baggage threat detection by proposing GANomaly \cite{akcay2018ganomaly}, an encoder-decoder-encoder network trained in an adversarial manner to recognize prohibited items within baggage X-ray scans. In another work, they proposed Skip-GANomaly \cite{akccay2019skip} which employs skip-connections in an encoder-decoder topology that not only gives better latent representations for detecting baggage threats but also reduces the overall computational complexity of GANomaly \cite{akcay2018ganomaly}.  

\begin{figure}[t]
    \centering
    \includegraphics[width=1\linewidth]{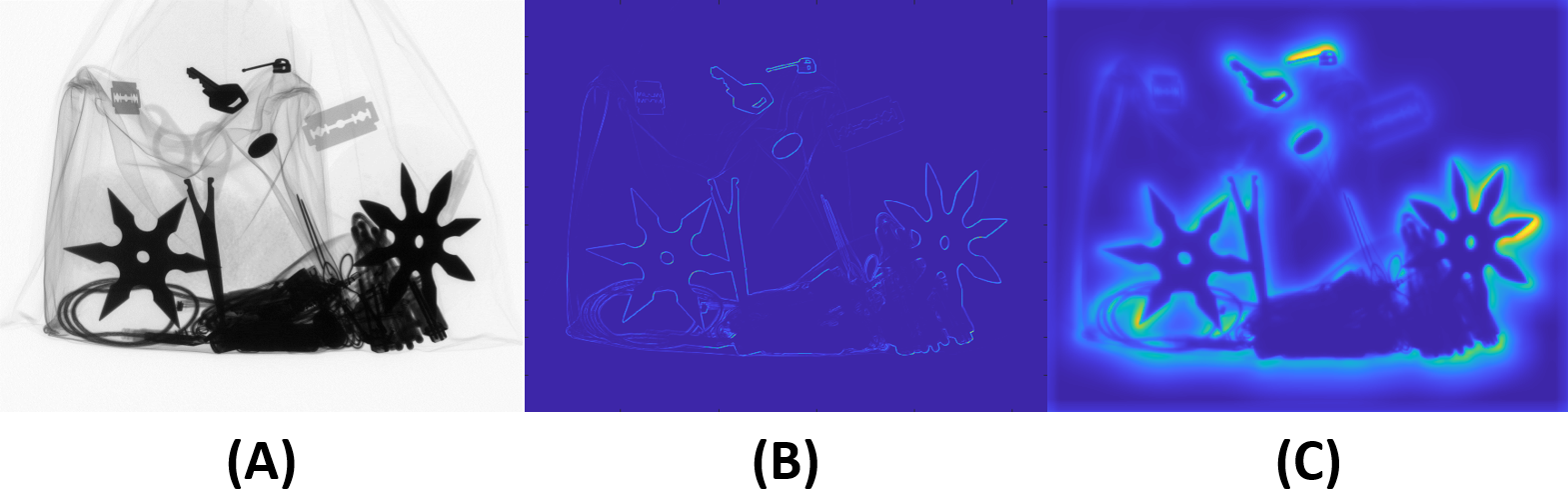}
    \caption{(A) An exemplar X-ray scan from the GDXray dataset \cite{mery2015gdxray}, (B) contour map obtained through the modified structure tensors in \cite{hassan2019} and \cite{Hassan2020ACCV}, (C) contour map obtained through proposed tensor pooling strategy.}
    \label{fig:fig1}
\end{figure}

\noindent The rest of the paper is organized as follows: Section \ref{sec:proposed} presents the proposed framework. Section \ref{sec:exp} describes the experimental setup.   Section \ref{sec:results} discusses the  results obtained with  three public baggage X-ray datasets.  Section \ref{sec:discussion} concludes the paper and enlists future directions.

\begin{figure*}[t]
    \centering
    \includegraphics[width=1\linewidth]{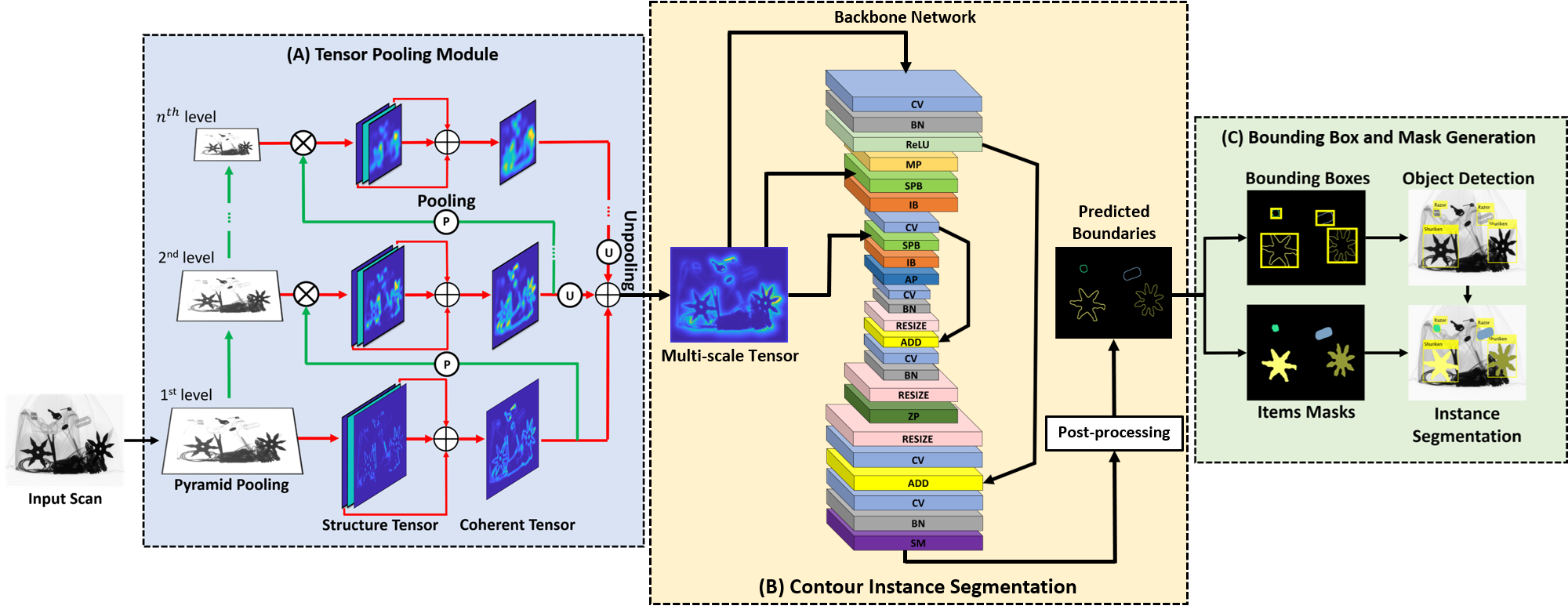}
    \caption{Block diagram of the proposed framework. The input scan is passed to the tensor pooling module to extract the tensor representations encoding the baggage items' contours at different orientations.  These representations are fused into a single multi-scale tensor and passed afterward to an asymmetric encoder-decoder backbone that segments and recognizes the contraband item's contours while suppressing the rest of the baggage content. For each detected contour, the corresponding bounding box and mask is generated to localize the detected contraband items. The abbreviations are CV: Convolution, BN: Batch Normalization, SPB: Shape Preserving Block, IB: Identity Block, MP: Max Pooling, AP: Average Pooling, ZP: Zero Padding, SM: Softmax.} 
    \label{fig:fig2}
\end{figure*}

\section{Proposed Approach } \label{sec:proposed}
\noindent The block diagram of the proposed framework is depicted in Figure \ref{fig:fig2}. The input scan is fed to the tensor pooling module (block A) to generate a multi-scale tensor representation, revealing the baggage content's transitional patterns at multiple predominant orientations and across various scales. Afterward, the multi-scale tensor is passed to the encoder-decoder backbone (block B), implementing the newly proposed contour maps-based instance segmentation. This block extracts the contours of the prohibited data while eliminating the irrelevant scan content. In the third stage (block C),  each extracted contour, reflecting the contraband item instance, is utilized in generating the respective mask and the bounding box for localization. In the subsequent sections, we present a detailed description of each module within the proposed framework.

\subsection{Tensor Pooling Module}
\noindent The tensor pooling module decomposes the input scan \RV{into $n$ levels of a pyramid. From each level of the pyramid,} the baggage content's transitional patterns are generated by analyzing their distribution of orientations within the associated image gradients.  In the proposed tensor pooling scheme, we highlight the transitional patterns in  $N$ image gradients (corresponding to  $N$ directions) by computing the following $N \times N$  block-structured  symmetric matrix \cite{hassan2019, Hassan2020ACCV}:

\begin{equation}
\begin{bmatrix}
  \phi * (\nabla^0 . \nabla^0) & \phi * (\nabla^1 . \nabla^0) & \cdots & \phi * (\nabla^{N-1} . \nabla^0) \\
  \phi * (\nabla^0 . \nabla^1) & \phi * (\nabla^1 . \nabla^1)  & \cdots & \phi * (\nabla^{N-1} . \nabla^1) \\
  \vdots & \vdots & \ddots & \vdots \\
  \phi * (\nabla^0 . \nabla^{N-1}) & \phi * (\nabla^1 . \nabla^{N-1}) & \cdots & \phi * (\nabla^{N-1} . \nabla^{N-1}) \\
\end{bmatrix}
\label{eq:eq3},
\end{equation}

\noindent
Each tensor ($\phi * (\nabla^k . \nabla^m)$) in the above block-matrix is an outer product of two image gradients and a smoothing filter $\phi$. Moreover, the orientation  ($\theta$), of the image gradient $\nabla^j$, is computed through: $\theta = \frac{2 \pi j}{N}$, where $j$ ranges from $0$ to $N-1$. Since the block-structured matrix in Eq. \ref{eq:eq3} is symmetric, we obtain $\frac{N(N-1)}{2}$ unique tensors. From this group, we derive the \RV{coherent tensor, reflecting the baggage items' predominant orientations. The coherent tensor is a single tensor representation generated by adding the most useful tensors out of the $\frac{N(N+1)}{2}$ unique tensor set. Here, it should be noted that these useful tensors are selected by ranking all the $\frac{N(N+1)}{2}$ unique tensors according to their norm.}  
 
\begin{figure}[htb]
    \centering
    \includegraphics[width=1\linewidth]{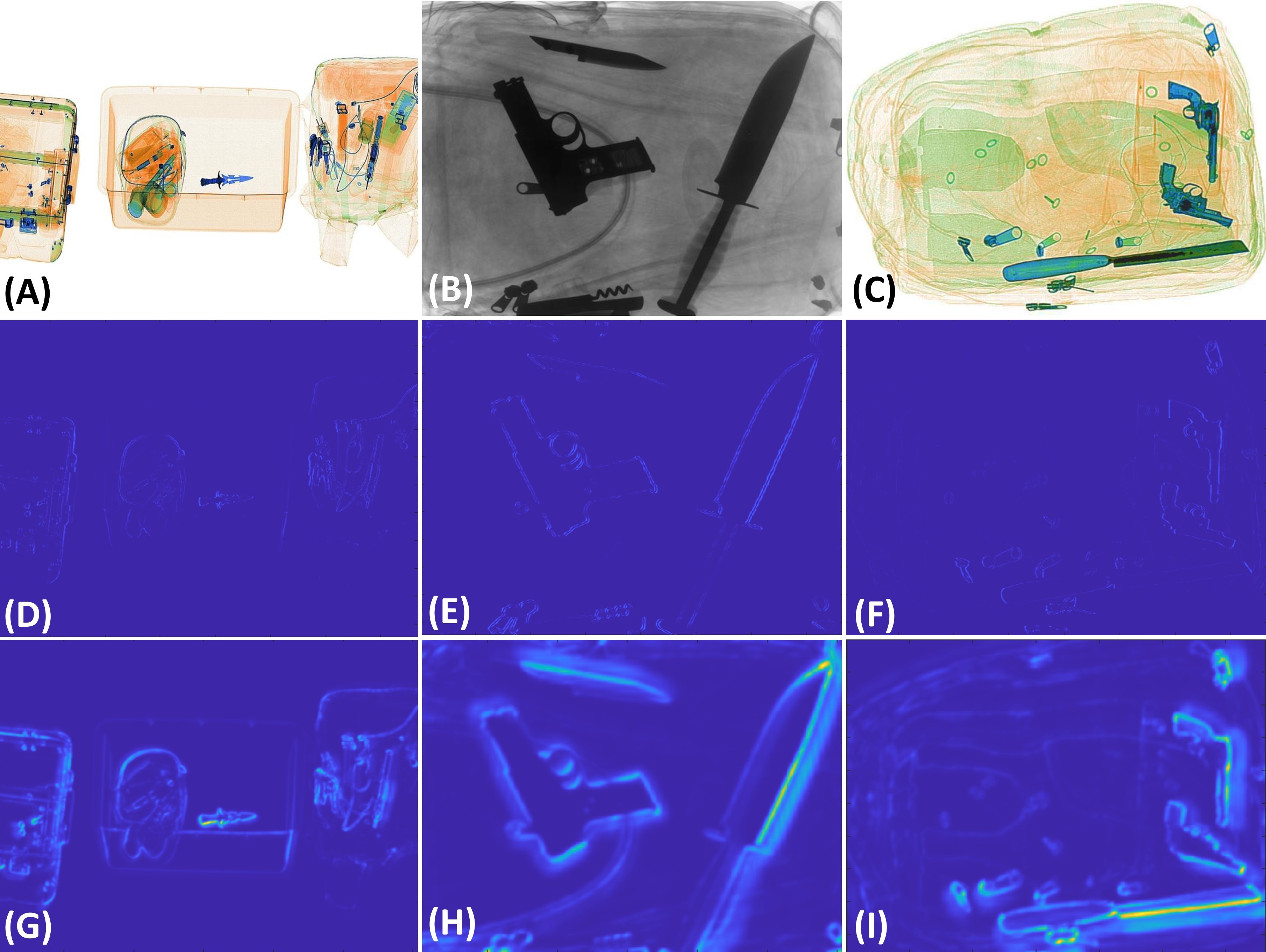}
    \caption{Difference between conventional structure tensors (used in \cite{hassan2019, Hassan2020ACCV}), and proposed multi-scale tensor approach. First row shows the original scans from OPIXray \cite{opixray}, GDXray \cite{mery2015gdxray}, and SIXray \cite{miao2019sixray} dataset. The second row shows the output for the conventional structure tensors \cite{hassan2019, Hassan2020ACCV}. The third row shows the output for the proposed tensor pooling module.}
    \label{fig:multiscale}
\end{figure}

\noindent Moreover, the coherent tensor also reveals the variations in the intensity of the cluttered baggage items, aiding in generating individual contours for each item.  However, this scheme analyzes only the intensity variations of the baggage items at a single scale, limiting the extraction of the objects having lower transitions with the background \cite{hassan2019, Hassan2020ACCV}. To address this limitation, we propose a multi-scale tensor fusing the  X-ray scan transitions from coarsest to finest levels so that each item, even having a low-intensity difference with the background, can be adequately highlighted.  
For example, see  the boundaries of:  the \textit{razor} in a multi-scale tensor representation in Figure \ref{fig:fig1} (C), the \textit{straight knife} in Figure 
\ref{fig:multiscale} (G), the two \textit{knives} and a \textit{gun} in Figure \ref{fig:multiscale} (H), and the two \textit{guns} and a \textit{knife} in Figure \ref{fig:multiscale} (I).
\begin{algorithm}[t]
\SetAlgoLined
\DontPrintSemicolon
\textbf{Input: } X-ray scan ($I$), Scaling Factor ($n$), Number of Orientations ($N$)

\textbf{Output: } Multi-scale Tensor ($M_t$)

[$r$, $c$] = size($I$)

Initialize $M_t$ (of size $r \times c$) with zeros 

Set $\eta=2$ // pyramid pooling factor

\For{$i=0$ to $n-1$}
{
    \eIf{$i$ is 0}
    {
        $\Im$ = ComputeTensors($I$, $N$) // $\Im$: Tensors
        
        $\Im_c$ = GetCoherentTensor($\Im$)
        
        $M_t$ = $M_t$ + $\Im_c$ 
    }
    {
        [$s$, $t$] = size($I$)
        
        \If{(min($s$, $t$) \% $\eta) \neq 0$ or min($s$, $t$) $< \eta$}       
        {
            break 
        }
        
        $I$ = Pool($I$, $\eta$) 
        
        $\Im_c$ = Pool($\Im_c$, $\eta$)  
        
        $I$ = $I \times \Im_c$ 

        $\Im$ = ComputeTensors($I$, $N$)  
        
        $\Im_c$ = GetCoherentTensor($\Im$)
        
        $M_t$ = $M_t$ + Unpool($\Im_c$, $\eta^i$)
    }
}
 \caption{Tensor Pooling Module}
 \label{algo:algo1}
\end{algorithm}

\noindent As mentioned earlier, the multi-scale tensors are computed through pyramid pooling (up to $n^{th}$ level).  At any $l^{th}$ level, (such that $2 \leq l \leq n$), we multiply, pixel-wise, the decomposed image with the transitions obtained at the previous ($l-1$) levels. In so doing, we ensure that the edges of the contraband items (procured earlier) are retained across each scale. The full procedure of the proposed tensor pooling module is depicted in Algorithm \ref{algo:algo1} and also shown in Figure \ref{fig:fig2}.
\begin{figure}[htb]
    \centering
    \includegraphics[width=1\linewidth]{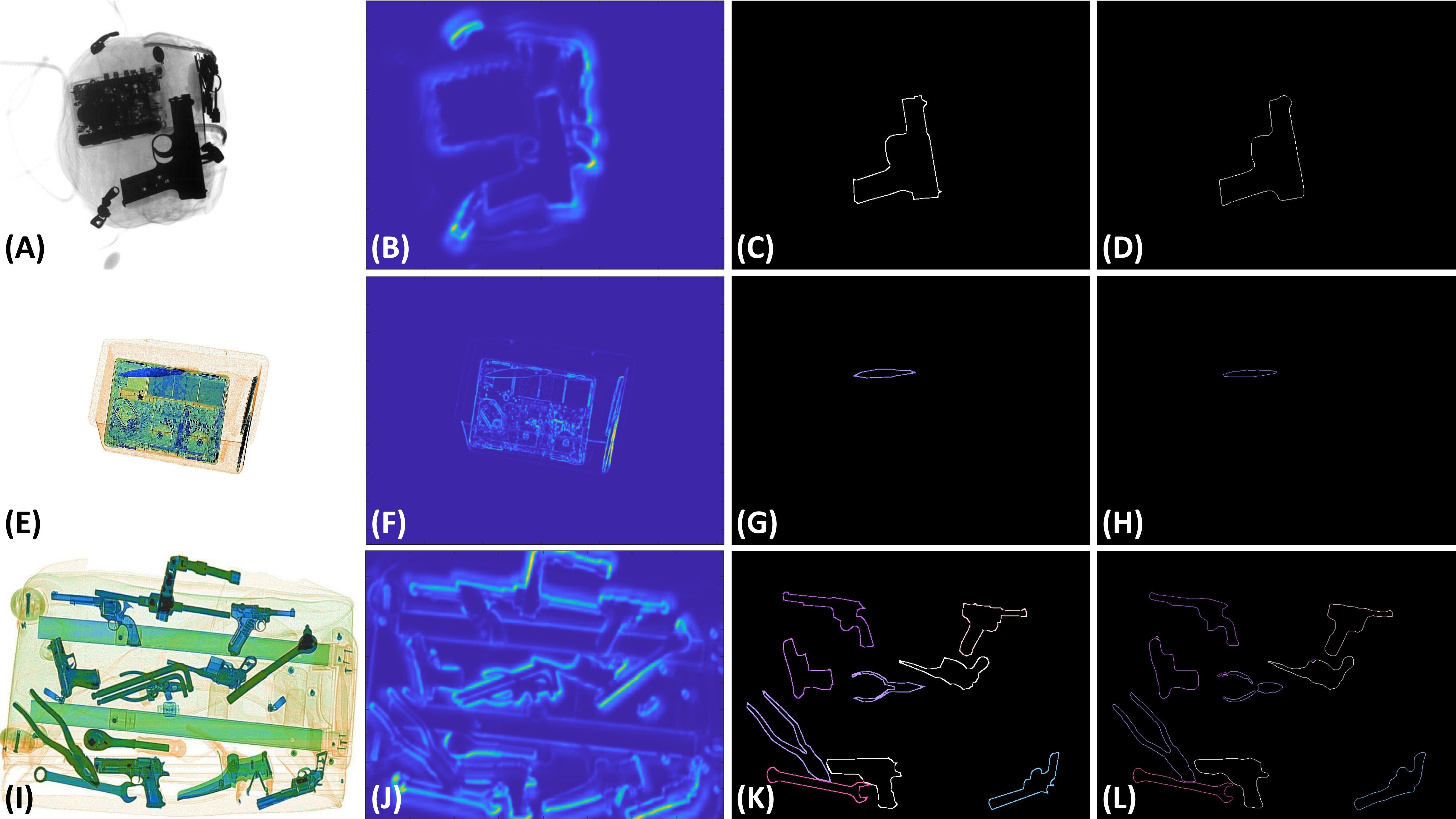}
    \caption{Contour instance segmentation from multi-scale tensors. The first column shows the original scans, the second column shows the multi-scale tensor representations, the third column shows the ground truths, and the fourth column shows the extracted contours of the contraband items.}
    \label{fig:contours}
\end{figure}

\noindent The multi-scale tensor is then passed to the proposed encoder-decoder model to extract the contours of the individual suspicious items. A detailed discussion about contour instance segmentation is presented in the subsequent section. 

\subsection {Contour Instance Segmentation}
\noindent The contour instance segmentation is performed through the proposed asymmetric encoder-decoder network, which assigns the pixels in the multi-scale tensors  to one of the following categories $\mathcal{C}_{k=1:\mathcal{M}+1}$ where $\mathcal{M}$ denotes the number of prohibited items' instances 
to which we add the class  \textit{background} which include background and irrelevant pixels (i.e., pixels belonging to a non-suspicious baggage content). 

\begin{figure}[t]
    \centering
    \includegraphics[width=1\linewidth]{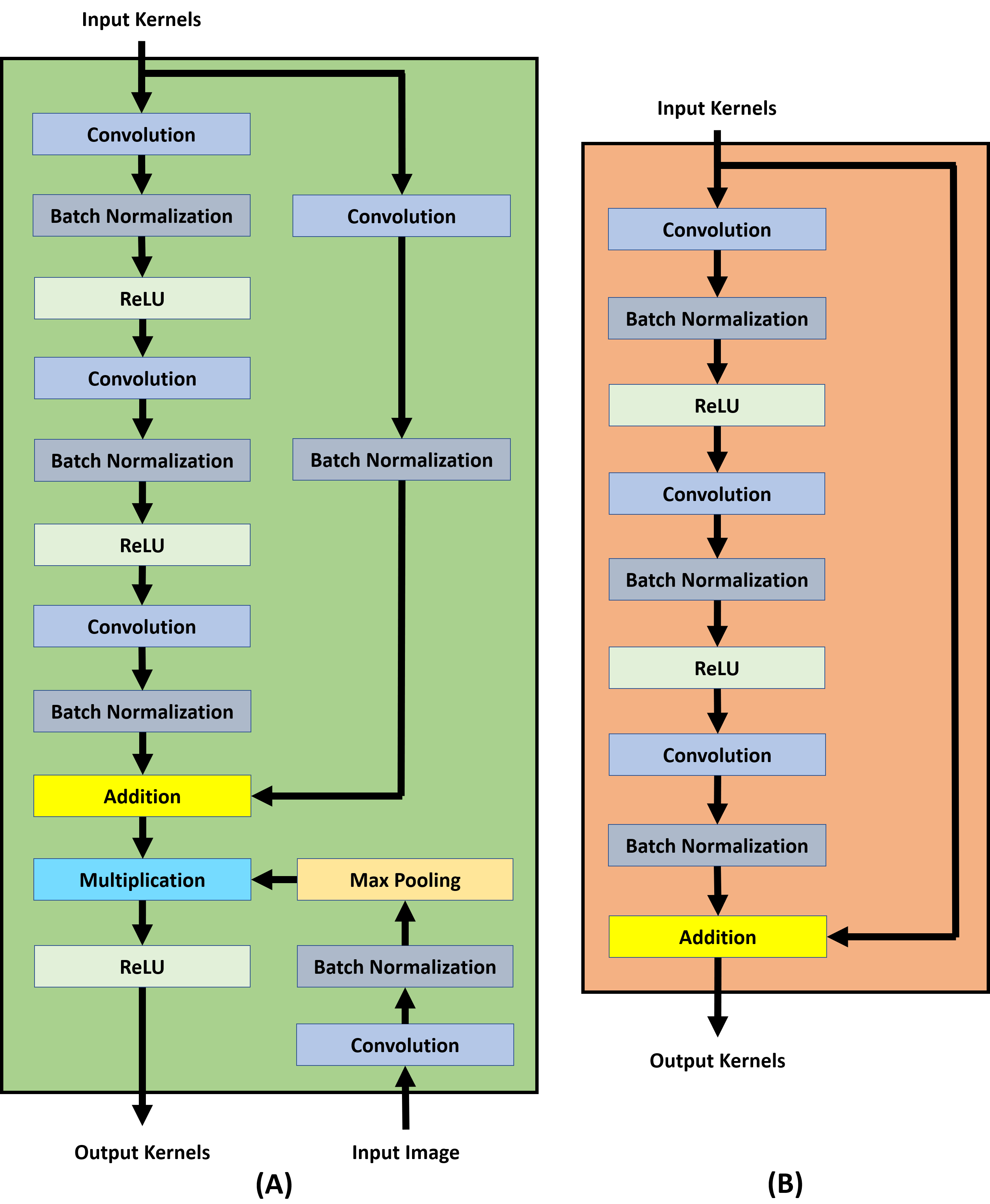}
    \caption{(A) Shape Preserving Block (SPB), (B) Identity Block (IB).}
    \label{fig:fig3}
\end{figure}

\noindent Furthermore, to differentiate between the contours of the normal and suspicious items, the custom shape-preserving (SPB) and identity blocks (IB) have been added within the encoder topology. The SPB, as depicted in Figures \ref{fig:fig2}  and \ref{fig:fig3} (A), integrates the multi-scale tensor map (after scaling) in the feature map extraction to enforce further the attention on prohibited items' outlines.  The IB (Figure \ref{fig:fig3}-A), inspired by ResNet architecture \cite{he2016deep}, acts as a residual block to emphasize the feature maps of the previous layer. 

\noindent Apart from this, the whole network encompasses one input, one zero-padding, 22 convolution, 20 batch normalization, 12 activation, four pooling, two multiply, six addition, three lambda (that implements the custom functions), and one reshape layer. Moreover, we use skip-connections (via addition) within the encoder-decoder to refine the extracted items' boundaries. The number of parameters within the network is 1,308,160, from which around 6,912 parameters are non-trainable. The detailed summary of the proposed model (including the architectural details of the SPB and IB blocks) is available in the source code repository\footnote{\label{note1}The source code of the proposed framework along with its complete documentation is available at \url{https://github.com/taimurhassan/tensorpooling}.}.

\subsection{Bounding Box and Mask Generation}
\noindent
After segmenting the contours, we perform morphological post-processing to remove tiny and isolated fragments. The obtained outlines contain both open and closed contours of the underlying suspicious items. The closed contours can directly lead towards generating the corresponding item's mask. For open contours, we join their endpoints and then derive their masks through morphological reconstruction. Afterward, we generate the items' bounding boxes from the masks as shown in Figure \ref{fig:fig2} (C).

\section{Experimental Setup} \label{sec:exp}
\noindent This section presents the details about the experimental protocols, datasets, and evaluation metrics which were in order to assess the proposed system's performance and compare it with state-of-the-art methods. 

\subsection{Datasets}
We validated the proposed framework on three different publicly available baggage X-ray datasets, namely, GDXray \cite{mery2015gdxray}, SIXray \cite{miao2019sixray}, and OPIXray \cite{opixray}. The detailed description of these datasets \RV{are presented below.} 

\subsubsection{GDXray}
GDXray \cite{mery2015gdxray} \RV{was} first introduced in 2015 and it contains 19,407 high-resolution grayscale X-ray scans. The dataset is primarily designed for the non-destructive testing purposes and the scans within GDXray \cite{mery2015gdxray} are arranged into five categories, i.e., \textit{welds}, \textit{baggage}, \textit{casting}, \textit{settings} and \textit{nature}. But \textit{baggage} is the only relevant group for this study and it contains 8,150 grayscale X-ray scans. Moreover, the dataset also presents the detailed annotations for the prohibited items such as \textit{shuriken}, \textit{knives}, \textit{guns}, and \textit{razors}. As per the dataset standard, 400 scans from GDXray \cite{mery2015gdxray} were used for training purposes, while the remaining scans were used for testing purposes.

\subsubsection{SIXray}
SIXray \cite{miao2019sixray} is a recently introduced large-scale security inspection X-ray dataset. It contains a total of 1,059,231 colored X-ray scans from which 8,929 scans are positive (containing prohibited items such as \textit{knives}, \textit{wrenches}, \textit{guns}, \textit{pliers}, \textit{hammer} and \textit{scissors} along with their ground truths), and 1,050,302 are negative (containing only the normal items). To validate the performance against class imbalance, the authors of the dataset presented three subset schemes of the dataset, namely, SIXray10, SIXray100, and SIXray1000 \cite{miao2019sixray}. Moreover, SIXray \cite{miao2019sixray} is also the largest and most challenging dataset (to date) designed to assess threat detection frameworks' performance towards screening extremely cluttered and highly imbalanced contraband data \cite{miao2019sixray, hassan2020Sensors}. As per the SIXray \cite{miao2019sixray} dataset standard,  we used 80\% scans for the training and the rest of 20\% for testing. 

\subsubsection{OPIXray}
OPIXray \cite{opixray} is the most recent baggage X-ray dataset (released publicly for the research community in 2020). It contains 8,885 colored X-ray scans. As per the dataset standard, out of these 8,885 scans, 7,109 are to be utilized for the training purposes, while the remaining 1,776 are to be used for testing purposes, to detect \textit{scissor}, \textit{straight knife}, \textit{multi-tool knife}, \textit{folding knife}, and \textit{utility knife}. Moreover, the dataset authors also quantified occlusion within the test scans into three levels, i.e., OP1, OP2, and OP3. OP1 indicates that the contraband items within the candidate scan contain no or slight occlusion, OP2 depicts a partial occlusion, while OP3 represents severe or full occlusion cases.

\RV{
\noindent We also want to highlight here that the resolution of the scans within each dataset varies significantly (except for OPIXray \cite{opixray}). For example, on GDXray \cite{mery2015gdxray}, the scan resolution varies as $2688 \times 2208$, $900 \times 1430$, $850 \times 850$ and $601 \times 1241$, etc. Similarly, on SIXray \cite{miao2019sixray}, the scan resolution varies as $681 \times 549 \times 3$, $801 \times 482 \times 3$, $649 \times 571 \times 3$, $1024 \times 640 \times 3$, and $675 \times 382 \times 3$, etc. But on OPIXray \cite{opixray}, the resolution of all the scans is 1225x954x3. 
In order to process all the scans with the proposed framework, we have re-sized them to the common resolution of $576 \times 768 \times 3$, which is extensively used in the recently published frameworks \cite{hassan2019, Hassan2020ACCV, hassan2020Sensors}.
}

\subsection{Training and Implementation Details}
\noindent 
The proposed framework \RV{was} developed using Python 3.7.4 with TensorFlow 2.2.0 and Keras APIs on a machine havingIntel Core i9-10940X@3.30 GHz CPU, 128 GB RAM and an NVIDIA Quadro RTX 6000 with cuDNN v7.5, and a CUDA Toolkit 10.1.243. Some utility functions are also implemented using MATLAB R2021a. Apart from this, the training on each dataset was conducted for a maximum of 50 epochs using ADADELTA \cite{Zeiler2012ADADELTA} as an optimizer (with the default learning and decay rate configurations) and a batch size of 4. Moreover, 10\% of the training samples from each dataset were used for the validation (after each epoch).  
For the loss function, we used the  focal loss \cite{retinanet}   expressed below:
\begin{equation}
l_f = -\frac{1}{b_s}\sum\limits_{i=0}^{b_s-1}\sum\limits_{j=0}^{c-1} \alpha(1-p(l_{i,j}))^\gamma t_{i,j}\log(p(l_{i,j}))
\label{eq:eq4}
\end{equation}
where $c$ represents the total number of classes, and $b_s$ denotes the batch size. $p(l_{i,j})$ denotes the predicted probability of the logit $l_{i,j}$ generated from $i^{th}$ training sample for the $j^{th}$ class, $t_{i,j}$ tells if the $i^{th}$ training sample actually belongs to the $j^{th}$ class or not, the term $\alpha(1-p(l_{i,j}))^\gamma$  represents the scaling factor that gives more weight to the imbalanced classes (in other words, it penalizes the network to give emphasize to the classes for which the network obtain low prediction scores). Through rigorous experiments, we empirically selected the optimal value of $\alpha$ and $\gamma$ as 0.25 and 2, respectively, as they result in faster learning for each dataset while simultaneously showing good resistance to the imbalanced data. \RV{Apart from this, architecturally, the kernel sizes within the proposed encoder-decoder backbone vary as 3x3 and 7x7, whereas the number of kernels varies as 64, 128, 256, 512, 1024, and 2048. Moreover, the pooling size within the network remained 2x2 across various network depths to perform the feature decomposition (at each depth) by the factor of 2. For more architectural and implementation details of the proposed framework, we refer the reader to the source code, which we have released publicly for the research community on GitHub\textsuperscript{\ref{note1}}.} 

\subsection{Evaluation Metrics}
In order to assess the proposed approach and compare it with the existing works, we used the following evaluation metrics:

\subsubsection{Intersection-over-Union}
Intersection-over-Union (IoU) tells how accurately the suspicious items have been extracted, and it is measured by checking the pixel-level overlap between the predictions and the ground truths. Mathematically, IoU is defined as:  
\begin{equation}
    IoU = \frac{T_p}{T_p+F_p+F_n}
    \label{eq:eq5},
\end{equation}
where $T_p$ are true positives (indicating that the pixels of the contraband items are correctly predicted w.r.t the ground truth), $F_p$ represents false positives (indicating that the background pixels are incorrectly classified as positives), and $F_n$ represents false negatives (meaning that the pixels of the contraband items are misclassified as background). Furthermore, we also calculated the mean IoU ($\mu$IoU) by taking an average of the IoU score for each contraband item class. 

\subsubsection{Dice Coefficient}
Apart from IoU scores, we also computed the dice coefficient (DC) scores to assess the proposed system's performance for extracting the contraband items. DC is calculated through:
 
\begin{equation}
    DC = \frac{2T_p}{2T_p+F_p+F_n}
    \label{eq:eq6},
\end{equation}

\noindent Compared to IoU, DC gives more weightage to the true positives (as evident from Eq. \ref{eq:eq6}). Moreover, the mean DC ($\mu$DC) is calculated by averaging DC scores for each category.

\subsubsection{Mean Average Precision}
The mean average precision (mAP) (in the proposed study) is computed by taking the mean of average precision (AP) score calculated for each contraband item class for the IoU threshold $\geq$ 0.5. Mathematically, mAP is expressed below:

\begin{equation}
    mAP = \sum_{i=0}^{n_c-1} AP(i)
    \label{eq:eq7},
\end{equation}

\noindent  where $n_c$ denotes the number of contraband items in each dataset. Here, we want to highlight that to achieve fair comparison with the state-of-the-art, we have used the original bounding box ground truths of each dataset for measuring the proposed framework's performance towards extracting the suspicious and illegal items. 

\section{Results} \label{sec:results}
\noindent In this section, we present the detailed results obtained with  GDXray \cite{mery2015gdxray},  SIXray \cite{miao2019sixray}, and OPIXray \cite{opixray} datasets. 
Before going into the experimental results, we present detailed ablation studies to determine the proposed framework's hyper-parameters. We also report a detailed comparison of the proposed encoder-decoder network with the popular segmentation models.

\subsection{Ablation Studies}
\noindent The ablation studies in this paper aim to determine the optimal values for 1) the number of orientations and scaling levels within the tensor pooling module and 2) the choice of the backbone model for performing the contour instance segmentation.

\begin{figure}[t]
    \centering
    \textbf{(A)} \includegraphics[width=0.4\linewidth,height=2.8cm]{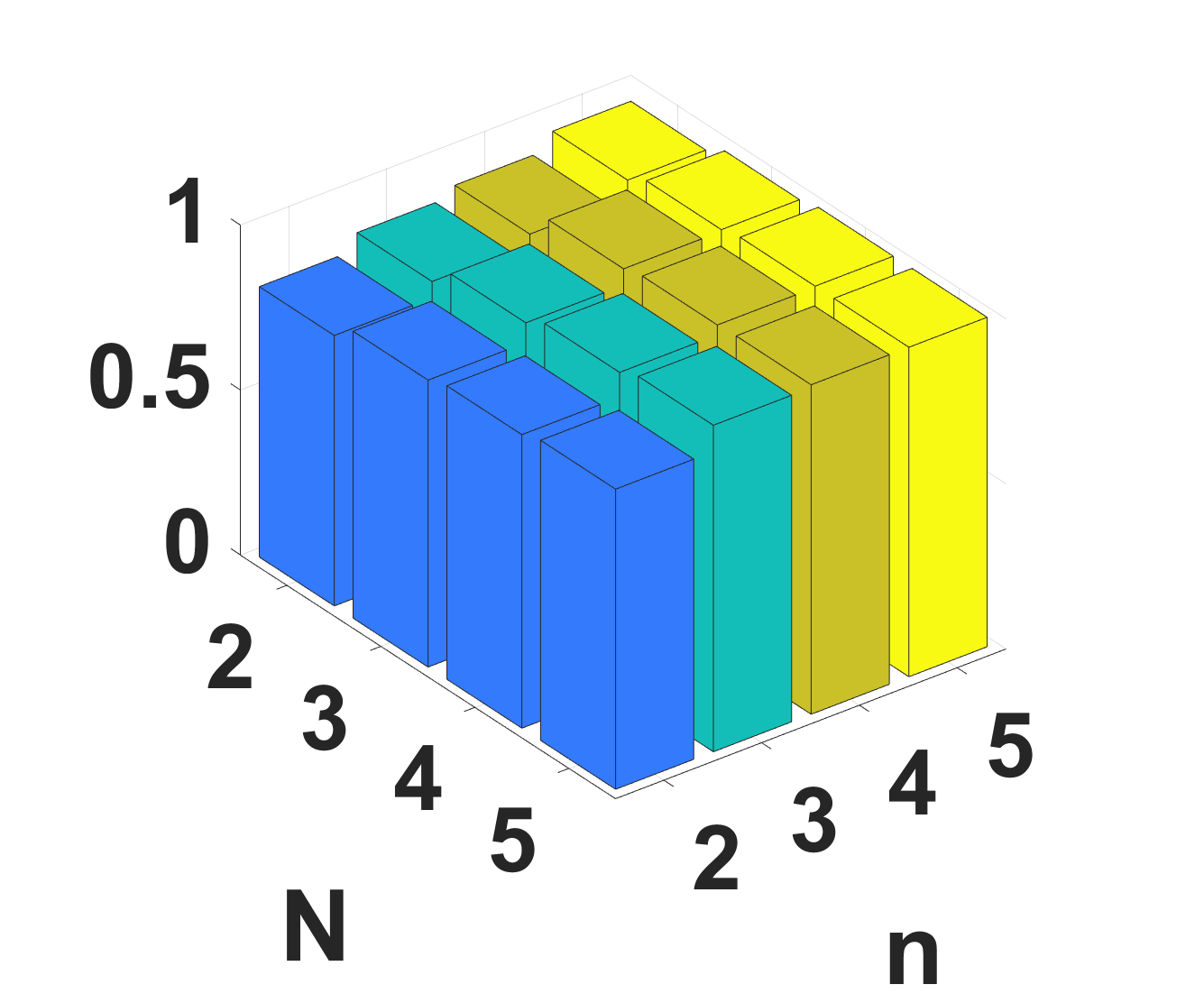}
    \textbf{(B)} \includegraphics[width=0.4\linewidth,height=2.8cm]{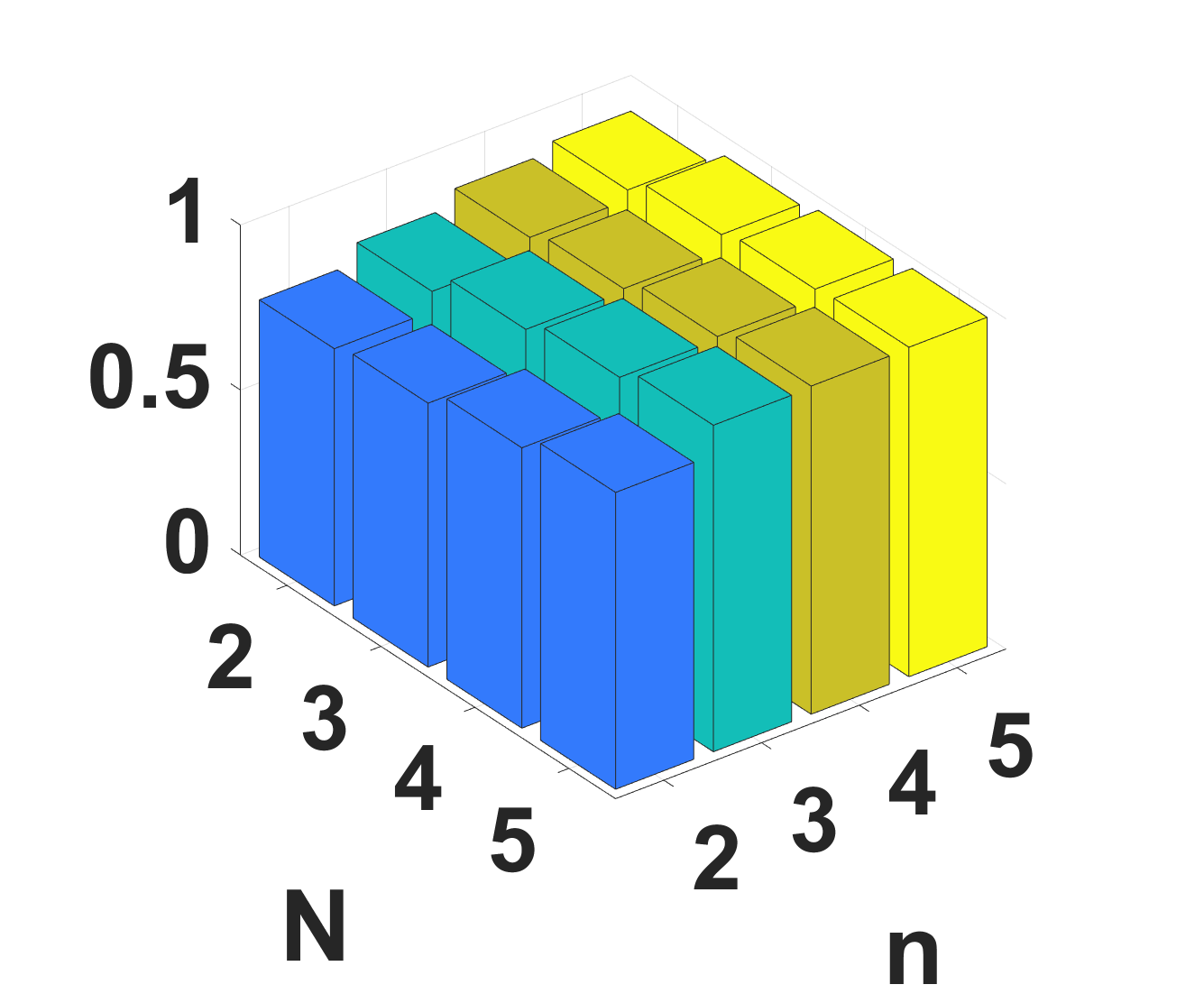}
   
    \textbf{(C)} \includegraphics[width=0.4\linewidth,height=2.8cm]{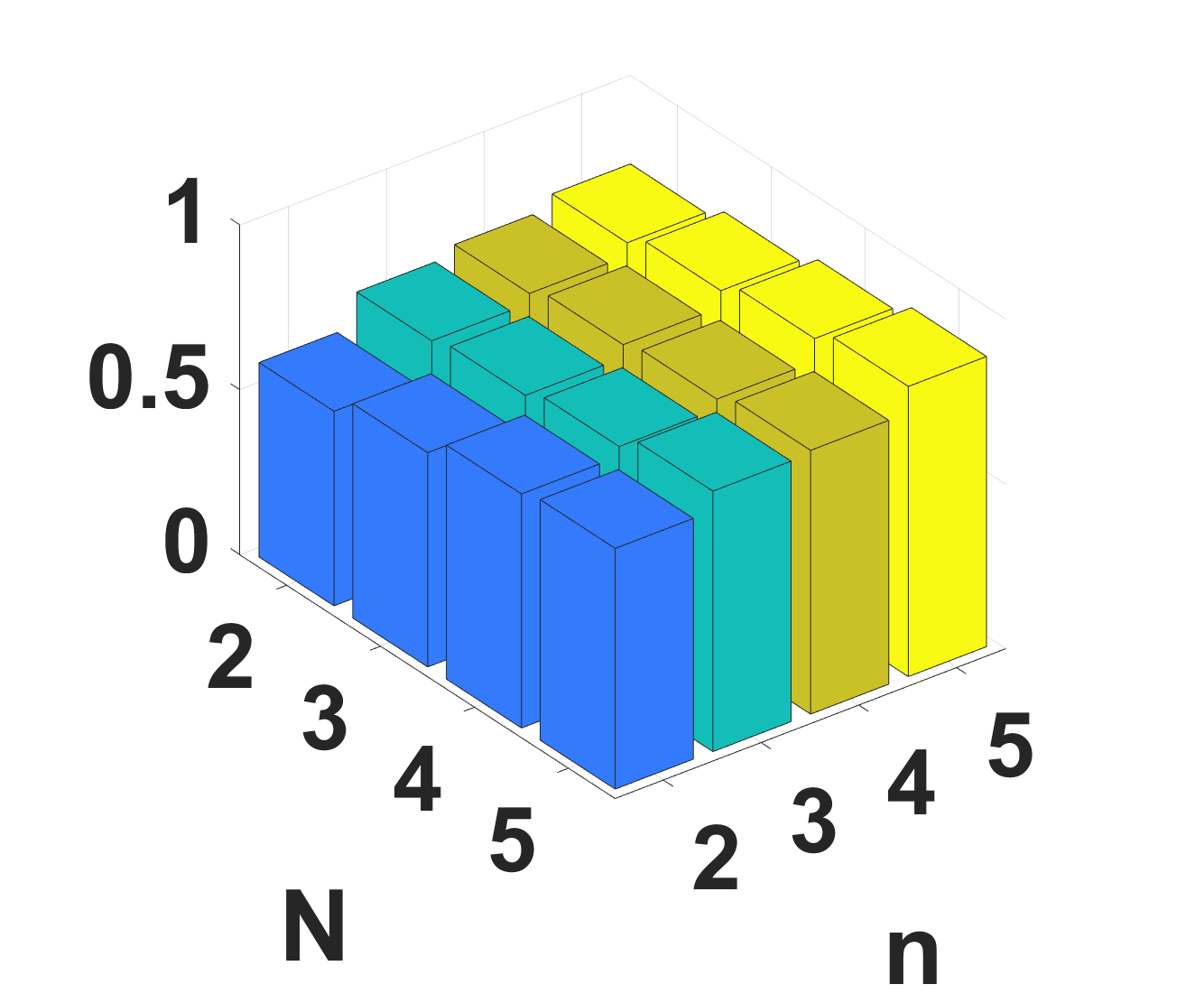}
    \textbf{(D)} \includegraphics[width=0.4\linewidth,height=2.8cm]{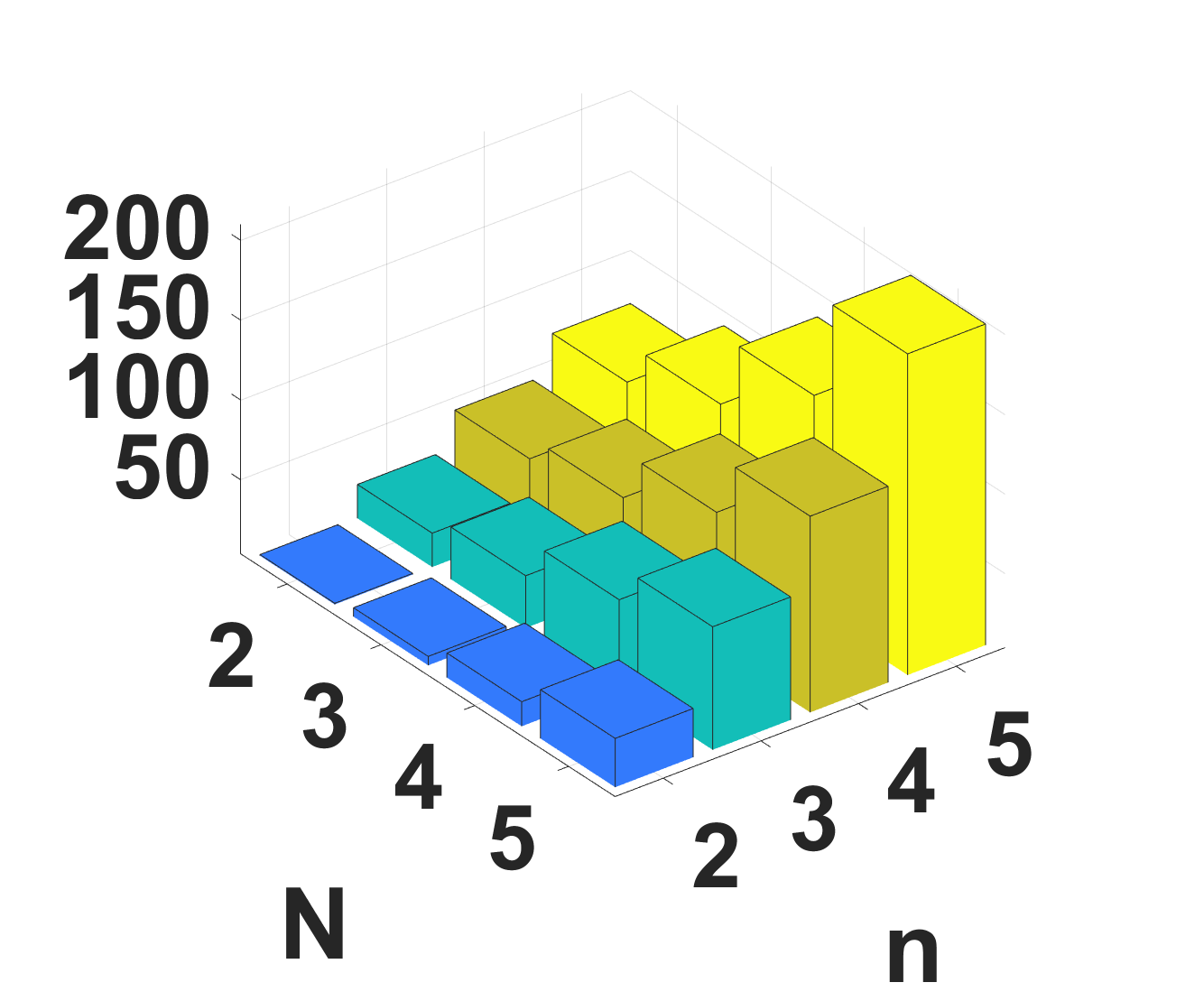}
    \textbf{(E)} \includegraphics[width=0.4\linewidth,height=2.8cm]{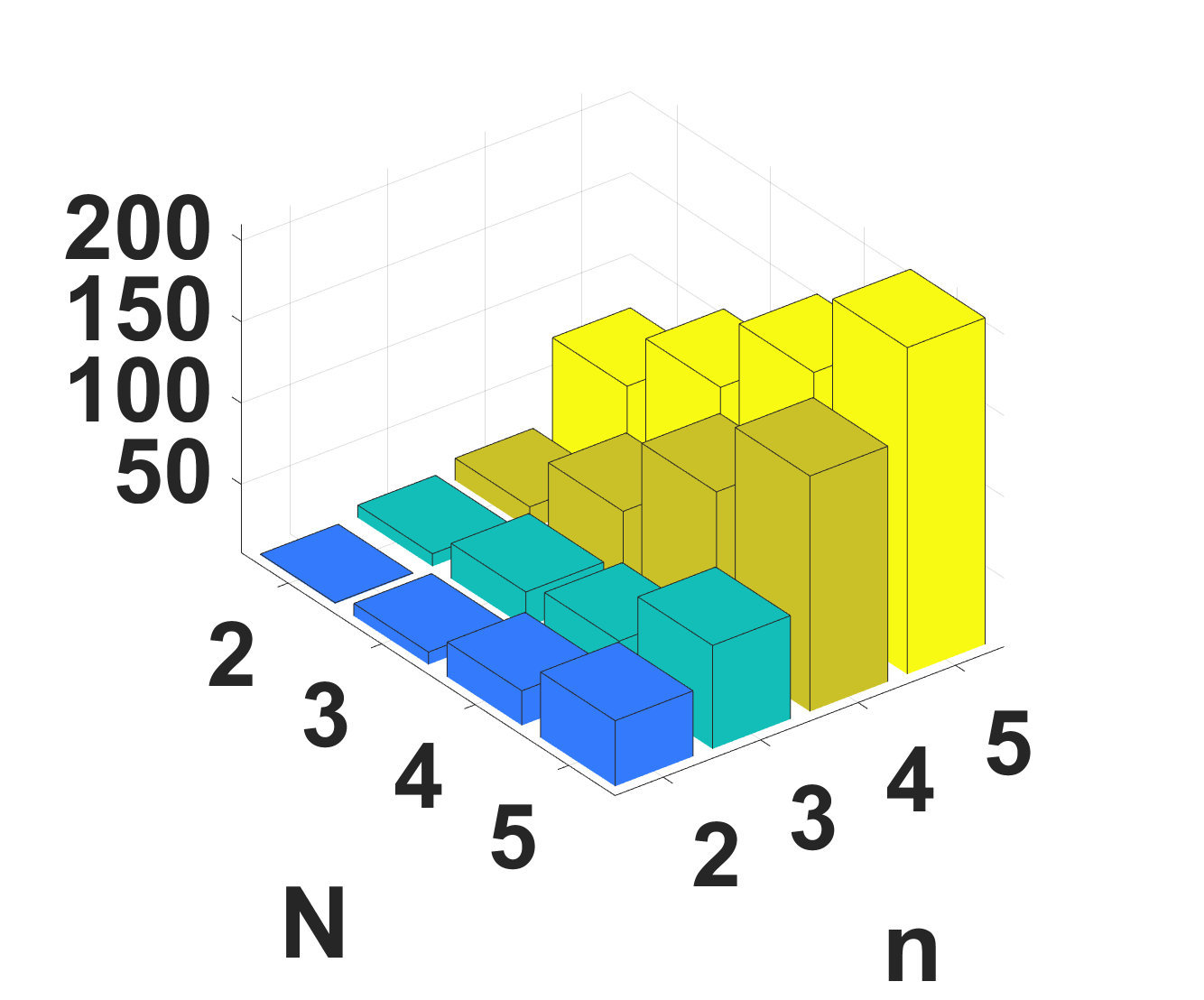}
    \textbf{(F)} \includegraphics[width=0.4\linewidth,height=2.8cm]{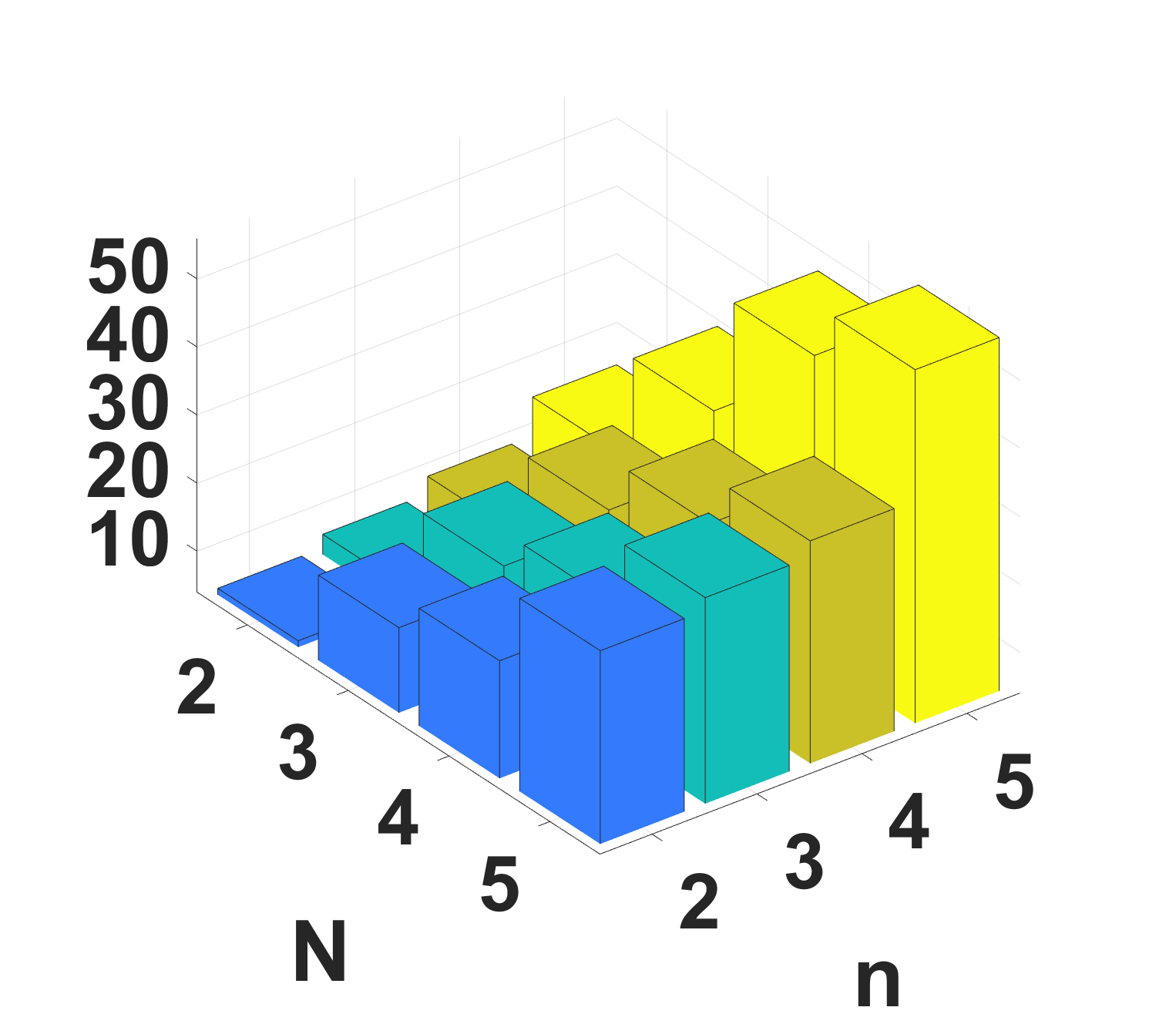}
    \caption{Detection performance of the proposed system in terms of mAP (A, B, C), and computational time in terms of seconds (D, E, F) obtained for GDXray \cite{mery2015gdxray}, SIXray \cite{miao2019sixray}, and OPIXray \cite{opixray} datasets, respectively.}
    \label{fig:fig33}

\end{figure}

\subsubsection{Number of Orientations and the Scaling Levels}
\noindent 
The tensor pooling module highlights the baggage content transitions in the image gradients oriented in $N$ directions and up to $n$ scaling levels. 
Increasing these parameters helps generate the best contour representation leading towards a more robust detection, but also incurs additional computational cost. As depicted  in Figure \ref{fig:fig33} (A), we can see that for GDXray dataset \cite{mery2015gdxray} with $N=2$, $n=2$, we obtain an mAP score of 0.82.  With  the combination $N=5$, $n=5$, we get \RV{16.54\%} improvements in the detection performance but at the expense of a \RV{97.71\%} increase in computational time (see Figure \ref{fig:fig33}-D). Similarly, on the SIXray dataset \cite{miao2019sixray}, we obtain \RV{18.36\%} improvements in the detection performance (by increasing $N$ and $n$) at the expense of \RV{95.88\%} in the computational time (see Figure \ref{fig:fig33} (B, E)). The same behavior is also noticed for OPIXray dataset \cite{opixray} in Figure \ref{fig:fig33} (C, F).    Considering all the combinations depicted in Figure \ref{fig:fig33}, we found that $N=4$  and $n=3$  provide the best trade-off between the detection and run-time performance across all three datasets.

\subsubsection{Choice of a Backbone Model}
\noindent 
The proposed backbone model has been specifically designed to segment the suspicious items' contours while discarding the normal baggage content. In this series of experiments, we compared the proposed asymmetric encoder-decoder model's performance with popular encoder-decoder, scene parsing, and fully convolutional networks. In terms of $\mu$DC and $\mu$IoU, we report the performance results in Table \ref{tab:tab2}.
We can observe that the proposed framework achieves the best extraction performance on OPIXray \cite{opixray} and SIXray \cite{miao2019sixray} dataset, leading the second-best UNet \cite{ronneberger2015unet} by \RV{2.34\%} and \RV{3.72\%}. On the GDXray \cite{mery2015gdxray}, however, it lags from the FCN-8 \cite{fcn8} and PSPNet \cite{zhao2017pyramid} by \RV{6.54\%} and \RV{5.91\%}, respectively. But as our model outperforms all the other architectures on the large-scale  SIXray \cite{miao2019sixray} and OPIXray \cite{opixray} datasets,  we chose it as a backbone for the rest of the experimentation.
\begin{table}[t]
    \centering
    \caption{Performance comparison of the proposed backbone network with PSPNet \cite{zhao2017pyramid}, UNet \cite{ronneberger2015unet} and FCN-8 \cite{fcn8} for recognizing the boundaries of the contraband items. The best and second-best performances are in bold and underline, respectively. Moreover, the abbreviations are: Met: Metric, Data: Dataset, GDX: GDXray \cite{mery2015gdxray}, SIX: SIXray \cite{miao2019sixray}, OPI: OPIXray \cite{opixray}.}
    \begin{tabular}{cccccc}
    \toprule
        Met & Data & Proposed & PSPNet & UNet & FCN-8 \\\hline
        $\mu$IoU & GDX & 0.4994 & \underline{0.5585} & 0.4921 & \textbf{0.5648} \\
        & SIX & \textbf{0.7072} & 0.5659 & \underline{0.6700} & 0.6613 \\
        & OPI & \textbf{0.7393} & 0.5645 & \underline{0.7159} & 0.5543 \\\hline
        
        $\mu$DC & GDX & 0.6661 & \underline{0.7167} & 0.6596 & \textbf{0.7219} \\
        & SIX & \textbf{0.8285} & 0.7227 & \underline{0.8024} & 0.7961 \\
        & OPI & \textbf{0.8501} & 0.7217 & \underline{0.8344} & 0.7132 \\
        \bottomrule
    \end{tabular}
    \label{tab:tab2}
\end{table}

\begin{table}[b]
    \centering
    \caption{Performance comparison between state-of-the-art baggage threat detection frameworks on GDXray (GDX), SIXray (SIX), and OPIXray (OPI) dataset in terms of mAP scores.  '-' indicates that the respective score is not computed. Moreover, the abbreviations are: Data: Dataset, GDX: GDXray \cite{mery2015gdxray}, SIX: SIXray \cite{miao2019sixray}, OPI: OPIXray \cite{opixray}, PF: Proposed Framework, and FD: FCOS \cite{fcos} + DOAM \cite{opixray}.}
    \begin{tabular}{cccccc}
    \toprule
        Data & Items & PF & CST & TST & FD \\\hline 
        GDX & Gun & \textbf{0.9872} & 0.9101 & \underline{0.9761} & - \\
        & Razor & \textbf{0.9691} & 0.8826 & \underline{0.9453} & - \\
        & Shuriken & 0.9735 & \textbf{0.9917} & \underline{0.9847} & -  \\
        & Knife & \underline{0.9820} & \textbf{0.9945} & 0.9632 & - \\
        & \color{blue}{mAP} & \textbf{0.9779} & 0.9281 & \underline{0.9672} & -  \\\hline
        
        SIX & Gun & \underline{0.9863} & \textbf{0.9911} & 0.9734 & -  \\
        & Knife & \textbf{0.9811} & 0.9347 & \underline{0.9681} & -  \\
        & Wrench & \underline{0.9882} & \textbf{0.9915} & 0.9421 & -  \\
        & Scissor & 0.9341 & \textbf{0.9938} & \underline{0.9348} & -  \\
        & Pliers & \underline{0.9619} & 0.9267 & \textbf{0.9573} & -  \\
        & Hammer & 0.9172 & \underline{0.9189} & \textbf{0.9342} & -  \\
        & \color{blue}{mAP} & \textbf{0.9614} & \underline{0.9595} & 0.9516 & -  \\\hline
        
        OPI & Folding & \underline{0.8528} & - & 0.8024 & \textbf{0.8671}  \\
        & Straight & \textbf{0.7649} & - & 0.5613 & \underline{0.6858}  \\
        & Scissor & 0.8803 & - &  \underline{0.8934} & \textbf{0.9023}  \\
        & Multi & \textbf{0.8941} & - &  0.7802 & \underline{0.8767} \\
        & Utility & \textbf{0.8062} & - & 0.7289 & \underline{0.7884}  \\
        & \color{blue}{mAP} & \textbf{0.8396} & - & 0.7532 & \underline{0.8241}  \\
    \bottomrule
    \end{tabular}
    \label{tab:tab3}
\end{table}

\subsection{Evaluation on GDXray Dataset}
\begin{figure}[t]
    \centering
    \includegraphics[width=1\linewidth]{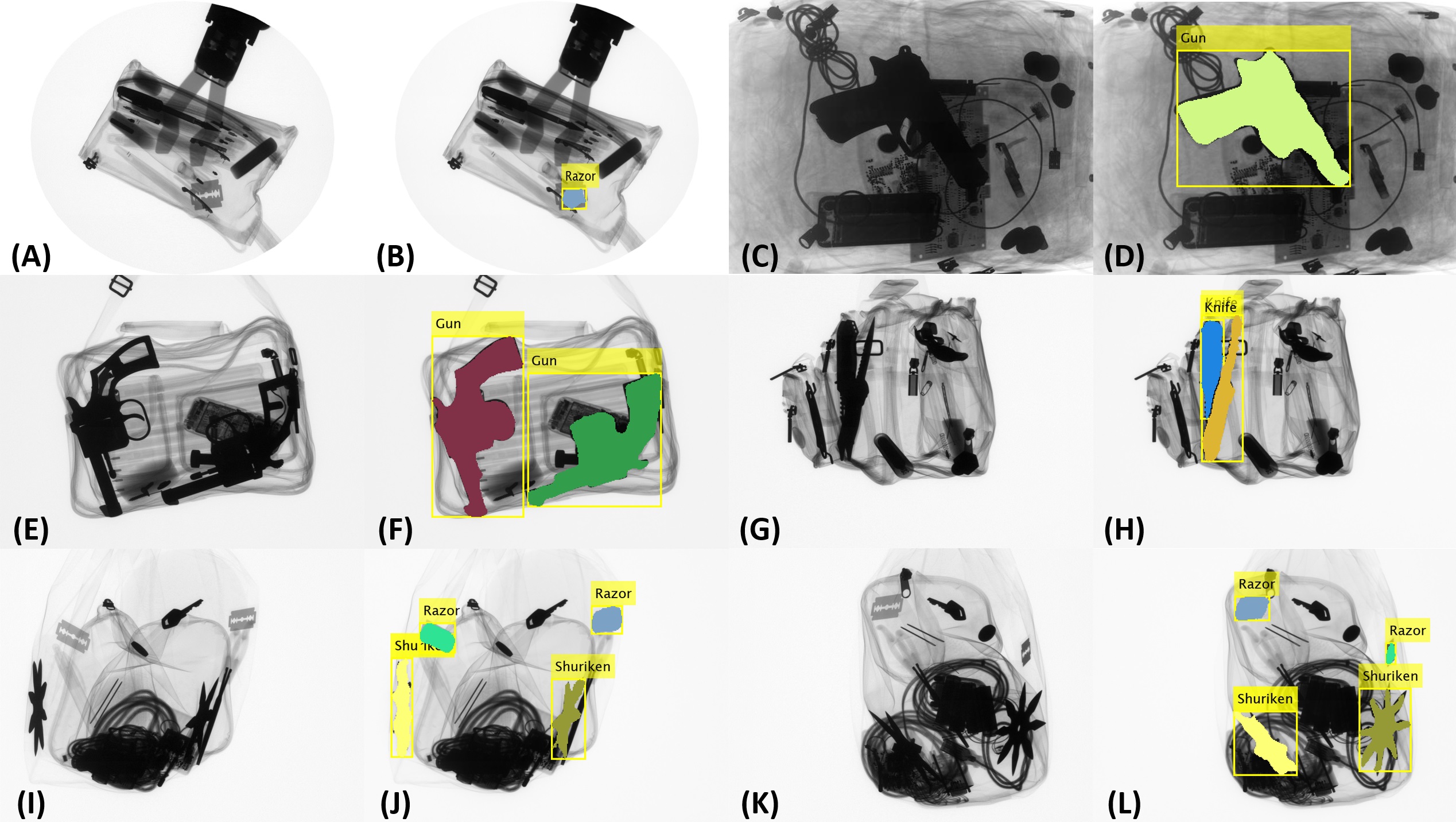}
    \caption{Qualitative evaluations of the proposed framework on GDXray \cite{mery2015gdxray} dataset. Please zoom-in for best visualization.}
    \label{fig:fig4}
\end{figure}

\noindent The performance of the proposed \RV{framework and of the state-of-the-art methods on the GDXray \cite{mery2015gdxray} dataset are reported} in Table \ref{tab:tab3}. We can observe here that the proposed framework outperforms the CST \cite{hassan2019} and the TST framework \cite{Hassan2020ACCV} by \RV{4.98\%} and \RV{1.07\%}, respectively. 
Furthermore, we wanted to highlight the fact that CST \cite{hassan2019} is only an object detection scheme, i.e., it can only localize the detected items but cannot generate their masks.   Masks are very important for the human observers in cross-verifying the baggage screening results (and identifying the false positives), especially from the cluttered and challenging grayscale scans.  
In  Figure \ref{fig:fig4},  we report some of the cluttered and challenging cases showcasing the effectiveness of the proposed framework
in extracting the overlapping contraband items. For example, see the extraction of merged \textit{knife} instances in (H), and the cluttered \textit{shuriken} in (J, L).  We can also appreciate how accurately the \textit{razors} have been extracted in (J, L).   Extracting such low contrast objects in the competitive CST framework requires suppressing first all the sharp transitions in an iterative fashion \cite{hassan2019}.

\subsection{Evaluations on SIXray Dataset}
\noindent 
The proposed framework has been evaluated on the whole SIXray dataset \cite{miao2019sixray} (containing 1,050,302 negative scans and 8,929 positive scans) and also on each of its subsets \cite{miao2019sixray}.  In Table \ref{tab:tab3}, we can observe that the proposed framework achieves an overall performance gain of \RV{0.190\%} and \RV{0.980\%} over CST \cite{hassan2019} and TST \cite{Hassan2020ACCV} framework, respectively.   
In Table \ref{tab:tab4}, we report the results obtained with each subset of the SIXray dataset \cite{miao2019sixray}, reflecting different imbalanced normal and prohibited item categories.  The results further confirm the superiority of the proposed framework against other state-of-the-art solutions, especially w.r.t the CHR \cite{miao2019sixray}, and \cite{gaus2019evaluating}. In addition to this, in an extremely challenging SIXray1000 subset,  we notice that the proposed framework leads the second-best TST framework \cite{Hassan2020ACCV} by \RV{3.22\%}, and CHR \cite{miao2019sixray} by \RV{44.36\%}. 

\begin{figure}[t]
    \centering
    \includegraphics[width=1\linewidth]{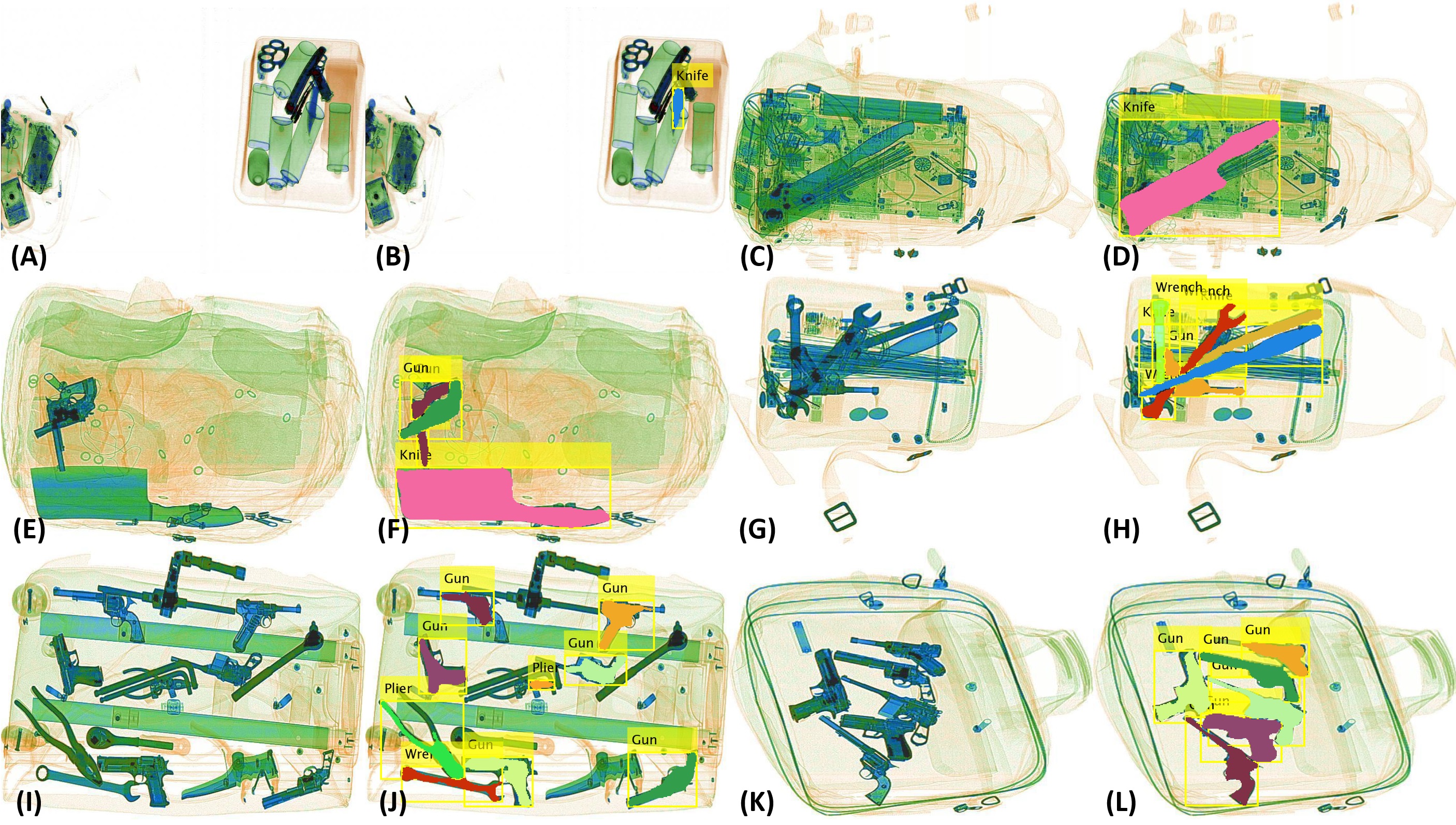}
    \caption{Qualitative evaluations of the proposed framework on SIXray \cite{miao2019sixray} dataset. Please zoom-in for a best visualization.}
    \label{fig:fig5}
\end{figure}

\noindent Apart from this, Figure \ref{fig:fig5} depicts the qualitative evaluations of the proposed framework on the SIXray \cite{miao2019sixray} dataset. In this figure, the first row shows examples containing one instance of the suspicious item, whereas the second and third rows show scans containing two or more instances of the suspicious items. 
Here, we can appreciate how accurately the proposed scheme has picked the cluttered \textit{knife} in (B). Moreover, we can also  observe the extracted \textit{chopper} (\textit{knife}) in (D) despite having  similar contrast with the background.   More examples such as (F, H, and J) demonstrate the proposed framework's capacity in picking the cluttered items from the SIXray dataset \cite{miao2019sixray}.

\begin{table}[t]
    \centering
    \caption{Performance comparison of proposed framework with state-of-the-art solutions on SIXray  subsets. For fair comparison, all models are evaluated using ResNet-50 \cite{he2016deep} as a backbone. Moreover, the abbreviations are: SIX-10: SIXray10 \cite{miao2019sixray}, SIX-100: SIXray100 \cite{miao2019sixray}, SIX-1k: SIXray1000 \cite{miao2019sixray}, and PF: Proposed Framework}
    \begin{tabular}{cccccc}
    \toprule
        Subset & PF & DTS & CHR & \cite{gaus2019evaluating} & TST\\\hline
        SIX-10 & \textbf{0.9793} & 0.8053 & 0.7794 & 0.8600 & \underline{0.9601}\\
        SIX-100 & \textbf{0.8951} & 0.6791 & 0.5787 & - & \underline{0.8749}\\
        SIX-1k & \textbf{0.8136} & 0.4527 & 0.3700 & - & \underline{0.7814}\\
        \bottomrule
    \end{tabular}
    \label{tab:tab4}
\end{table}

\subsection{Evaluations on OPIXray Dataset}
\noindent 
The performance evaluation of the proposed framework on OPIXray dataset \cite{opixray} is reported in Table \ref{tab:tab3}. We can observe here that the proposed system achieves an overall mAP score of 0.8396, outperforming the second-best DOAM framework \cite{opixray} (driven via FCOS \cite{fcos}) by \RV{1.55\%}.  Here, although the performance of both frameworks is identical, we still achieve a significant lead of \RV{7.91\%} over the DOAM \cite{opixray} for extracting the \textit{straight knives}.

\noindent  Concerning the level of occlusion (as aforementioned, OPIXray \cite{opixray} splits the test data into three subsets, OP1, OP2, OP3,  according to the level of occlusion), we can see in Table \ref{tab:tab5}  that the proposed framework achieves the best performance at each occlusion level as compared to the second-best DOAM \cite{opixray} framework driven by the single-shot detector (SSD) \cite{Liu2016SSD}. 

\begin{figure}[t]
    \centering
    \includegraphics[width=1\linewidth]{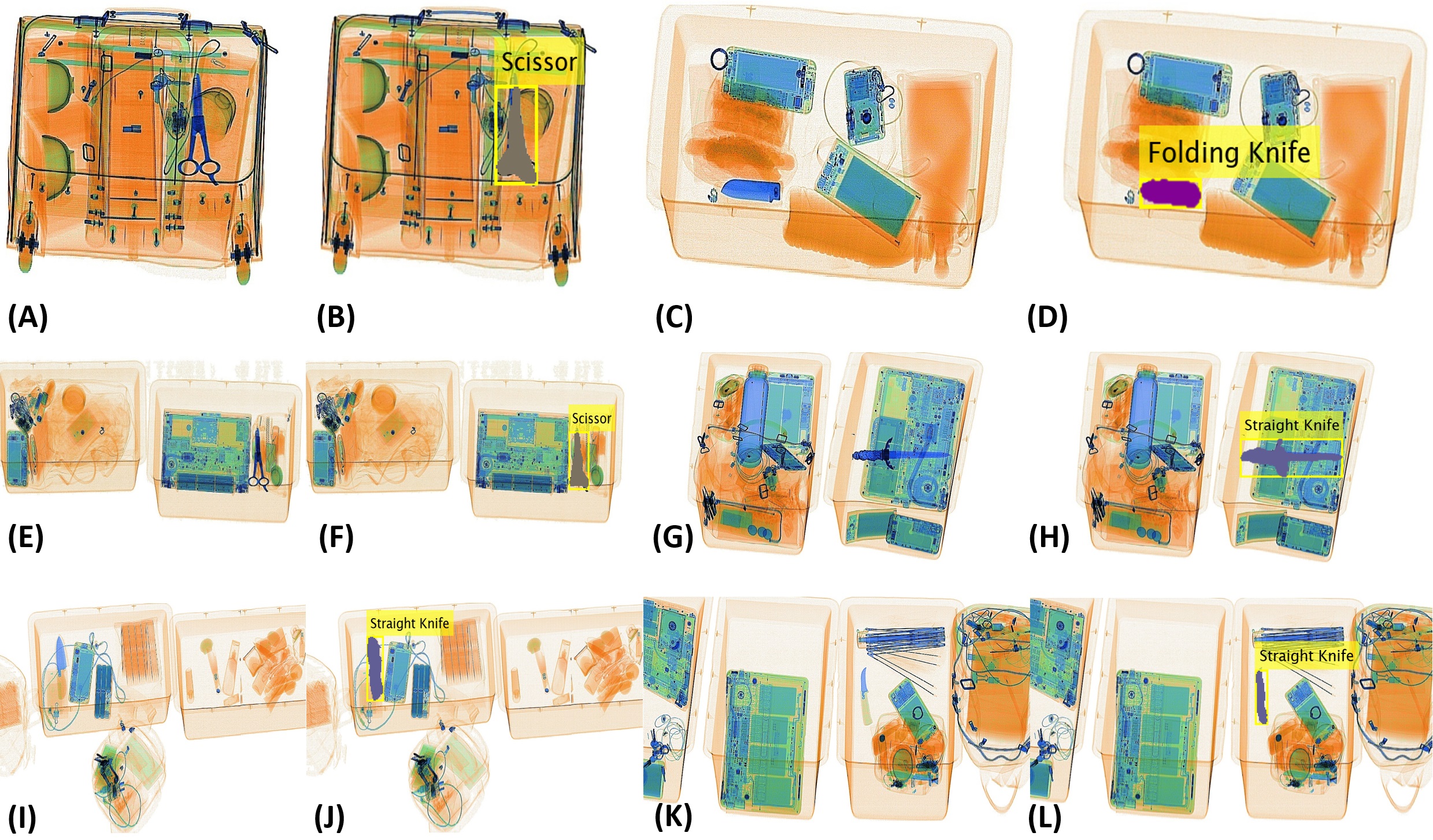}
    \caption{Qualitative evaluations of the proposed framework on OPIXray \cite{opixray} dataset. Please zoom-in for a best visualization.}
    \label{fig:fig6}
\end{figure}

\noindent  Figure \ref{fig:fig6} reports some qualitative evaluation, where we can appreciate the recognition of the cluttered \textit{scissor} (e.g. see B and F), and overlapping \textit{straight knife} (in H). We can also notice the detection of the partially occluded \textit{folding and straight knife} in (D) and (J). 

\subsection{Failure Cases}
\begin{figure}[t]
    \centering
    \includegraphics[width=1\linewidth]{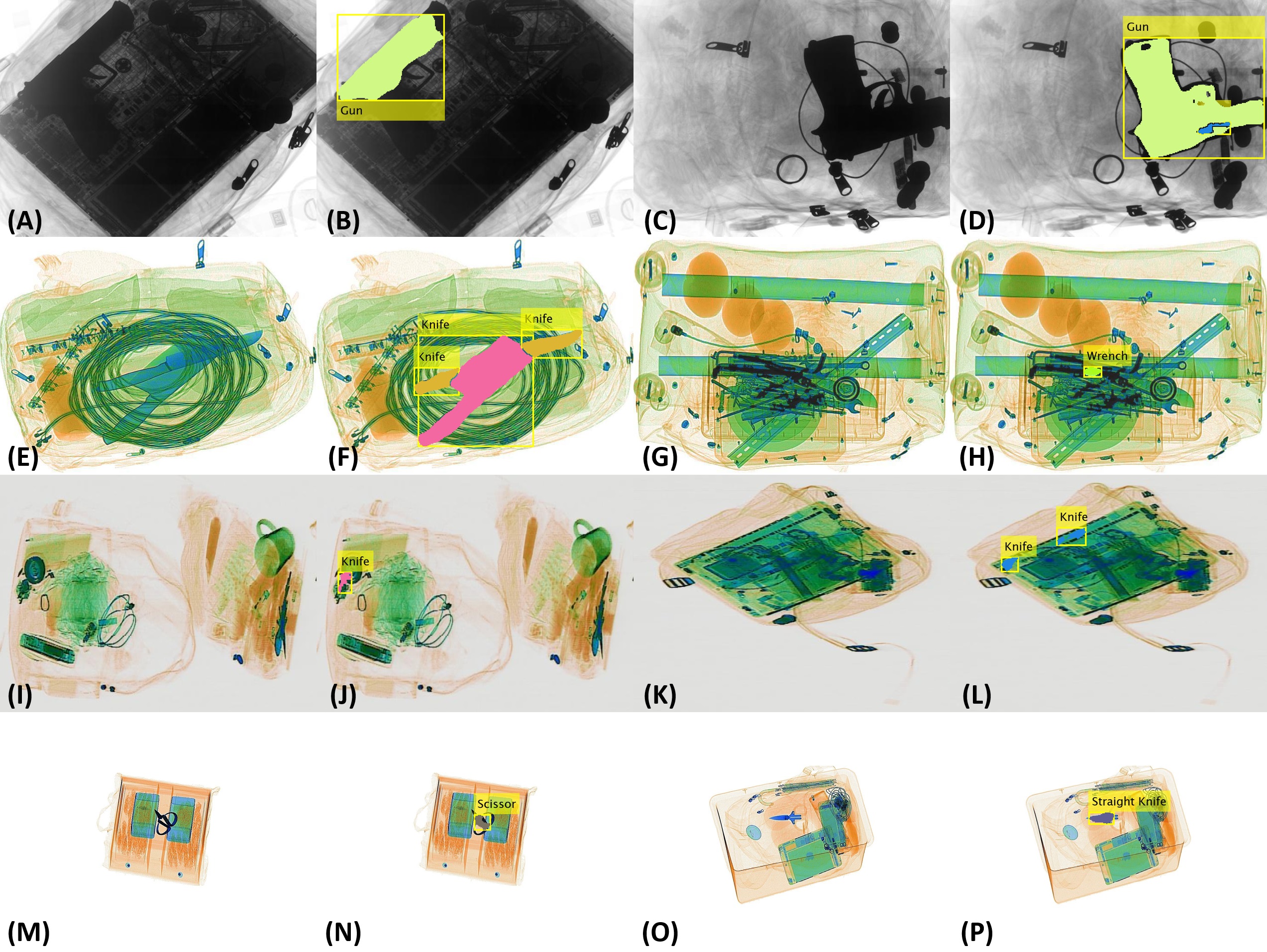}
    \caption{Failure cases from GDXray \cite{mery2015gdxray}, SIXray \cite{miao2019sixray}, and OPIXray \cite{opixray} dataset.}
    \label{fig:fig8}
\end{figure}

\noindent In Figure \ref{fig:fig8}, we report examples of failure cases encountered during the testing.   In cases (B, H, N, and P), we can see that the proposed framework could not pick-up the whole regions of the contraband items, even though the items were detected correctly. However, such cases are observed in highly occluded scans such as (A and G), where it is difficult, even for a human observer, to distinguish the items' regions properly. 
The second type of failure corresponds to the pixels misclassification as shown in (D) where some of the \textit{gun}'s pixels have been misclassified as \textit{knife}.  We can address these scenarios through post-processing steps like blob removal and region filling. 
The third failure case relates to the proposed framework's inability to generate a single bounding box for the same item. Such a case is depicted in (F), where two bounding boxes were generated for the single orange \textit{knife} item.  One possible remedy here is to generate the bounding boxes based upon the minimum and maximum mask value in both image dimensions for each label.  Another type of failure is shown in (J) and (L). Here, the scans contain only normal baggage content, but some pixels occupying tiny regions have been misclassified as false positive (i.e., \textit{knife}). We can also address this kind of failure through blob removal scheme. 

\noindent Examining the failure cases' statistical distributions,  we found a majority of 86.09\% cases belonging to the curable categories  (i.e., second, third, and fourth), meaning that the proposed framework's performance can be further improved using the post-processing techniques mentioned above.

\begin{table}[t]
    \centering
    \caption{Performance comparison of proposed framework with DOAM \cite{opixray} (backboned through SSD \cite{Liu2016SSD}) on different occlusion levels of OPIXray \cite{opixray} dataset.} 
    \begin{tabular}{cccc}
    \toprule
        Method & OP1 & OP2 & OP3 \\ \hline
        Proposed & \textbf{0.7946} & \textbf{0.7382} & \textbf{0.7291} \\
        DOAM + SSD \cite{opixray} & \underline{0.7787} & \underline{0.7245} & \underline{0.7078} \\
        SSD \cite{Liu2016SSD} & 0.7545 & 0.6954 & 0.6630\\
        \bottomrule
    \end{tabular}
    \label{tab:tab5}
\end{table}

\section{Conclusion} \label{sec:discussion}
\noindent 
In this work, we proposed a novel contour-driven approach for detecting cluttered and occluded contraband items (and their instances) within the baggage X-ray scans, hypothesizing that contours are the most robust cues given the lack of texture in the X-ray imagery. We concretized this original approach through a tensor pooling module, producing multi-scale tensor maps highlighting the items' outlines within the X-ray scans and an instance segmentation model acting on this representation. 
 We validated our approach on three publicly available datasets encompassing gray-level and colored scans and showcased its overall superiority over competitive frameworks in various aspects. For instance, the proposed framework outperforms the state-of-the-art methods \cite{opixray,hassan2019,Hassan2020ACCV,hassan2020Sensors} by \RV{1.07\%}, \RV{0.190\%}, and \RV{1.55\%} on GDXray \cite{mery2015gdxray}, SIXray \cite{miao2019sixray}, and OPIXray \cite{opixray} dataset, respectively. Furthermore, on each SIXray subsets (i.e., SIXray10, SIXray100, SIXray1000) \cite{miao2019sixray}, the proposed framework leads the state-of-the-art by \RV{1.92\%}, \RV{2.02\%}, and \RV{3.22\%}, respectively.
 
\noindent In future, we aim to apply the proposed framework to recognize 3D printed contraband items from the X-ray scans. Such items exhibit poor visibility in the X-ray scans because of their organic material, making them an enticing and challenging case to investigate and address.

\section*{Conflicts of Interest Declarations}

\noindent \textbf{Funding}: This work is supported by a research fund from ADEK (Grant Number: AARE19-156) and Khalifa University (Grant Number: CIRA-2019-047). 

\noindent \textbf{Conflict of Interest}: The authors have no conflicts of interest to declare that are relevant to this article.

\noindent \textbf{Financial and Non-Financial interests}: All the authors declare that they have no financial or non-financial interests to disclose for this article.

\noindent \textbf{Employment}: The authors conducted this research during their employment in the following institutes:
\begin{itemize}
\item T. Hassan (Khalifa University, UAE), 
\item S. Ak\c{c}ay (Durham University, UK), 
\item M. Bennamoun (The University of Western Australia, Australia), 
\item S. Khan (Mohamed bin Zayed University of Artificial Intelligence, UAE), and
\item N. Werghi (Khalifa University, UAE).
\end{itemize}

\noindent \textbf{Ethics Approval}: All the authors declare that no prior ethical approval was required from their institutes to conduct this research.

\noindent \textbf{Consent for Participate and Publication}: All the authors declare that no prior consent was needed to disseminate this article as there were no human (or animal) participants involved in this research.  

\noindent \textbf{Availability of Data and Material}: All the datasets that have been used in this article are publicly available.

\noindent \textbf{Code Availability}: The source code of the proposed framework is released publicly on GitHub\textsuperscript{\ref{note1}}.

\noindent \textbf{Authors' Contributions}: 
T. Hassan formulated the idea, wrote the manuscript, and performed the experiments. 
S. Ak\c{c}ay improved the initial design of the framework and contributed to manuscript writing. 
M. Bennamoun co-supervised the whole research, reviewed the manuscript and experiments. 
S. Khan reviewed the manuscript, experiments and improved the manuscript writing. 
N. Werghi supervised the whole research, contributed to manuscript writing, and reviewed the experimentation. 

\small

\end{document}